\documentclass[twoside,11pt]{article}

\usepackage{jmlr2e}
\usepackage{amsmath}
\usepackage{dsfont}
\usepackage[]{algorithm2e}
\usepackage{graphicx}
\usepackage{booktabs}
\usepackage{enumerate}
\usepackage{color}

\newcommand{\ve}[1]{{\boldsymbol{#1}}} 
\newcommand{\tr}{\mbox{tr}} 
\newcommand{\ex}{\mathds{E}} 
\newcommand{\va}{\mathds{V}} 
\newcommand{\N}{\mathcal{N}} 
\newcommand{\ord}{\mathcal{O}} 
\newcommand{\diag}{\mbox{diag}} 

\jmlrheading{?}{2017}{??--??}{04/17}{??/??}{Hildo Bijl, Thomas B. Sch\"on, Jan-Willem van Wingerden, Michel Verhaegen}

\ShortHeadings{System Identification through Online Sparse Gaussian Process Regression with Input Noise}{Bijl}
\firstpageno{1}

\begin{document}

\title{System Identification through Online Sparse Gaussian Process Regression with Input Noise}

\author{\name Hildo Bijl \email h.j.bijl@tudelft.nl \\
       \addr Delft Center for Systems and Control \\
       Delft University of Technology\\
       Delft, The Netherlands
       \AND
       \name Thomas B. Sch\"on \email thomas.schon@it.uu.se \\
       \addr Department of Information Technology \\
       Uppsala University \\
       Uppsala, Sweden
       \AND
       \name Jan-Willem van Wingerden \email j.w.vanwingerden@tudelft.nl \\
       \addr Delft Center for Systems and Control \\
       Delft University of Technology\\
       Delft, The Netherlands
       \AND
       \name Michel Verhaegen \email m.verhaegen@tudelft.nl \\
       \addr Delft Center for Systems and Control \\
       Delft University of Technology\\
       Delft, The Netherlands
			}

\editor{??}

\maketitle

\begin{abstract}
  There has been a growing interest in using non-parametric regression methods like Gaussian Process (GP) regression for system identification. GP regression does traditionally have three important downsides: (1) it is computationally intensive, (2) it cannot efficiently implement newly obtained measurements online, and (3) it cannot deal with stochastic (noisy) input points. In this paper we present an algorithm tackling all these three issues simultaneously. The resulting Sparse Online Noisy Input GP (SONIG) regression algorithm can incorporate new noisy measurements in constant runtime. A comparison has shown that it is more accurate than similar existing regression algorithms. When applied to nonlinear black-box system modeling, its performance is competitive with existing nonlinear ARX models.
\end{abstract}


\section{Introduction}

The Gaussian Process (GP)~\citep{GPBook} has established itself as a standard model for nonlinear functions. It offers a representation that is non-parametric and probabilistic. The \emph{non-parametric} nature of the GP means that it does not rely on any particular parametric functional form to be postulated. The fact that the GP is a \emph{probabilistic}  model means that it is able to take uncertainty into account in every aspect of the model.

The nonlinear system identification problem amounts to learning a nonlinear mathematical model based on data that is observed from a dynamical phenomenon under study. Recently there has been a growing interest of using the GP to this end and it has in fact allowed researchers to successfully revisit the linear system identification problem and establish new and significantly improved results on the impulse estimation problem~\citep{PillonettoDN:2010,PillonettoCDN:2011,ChenOL:2012}. There are also older results on nonlinear ARX type models~\citep{KocijanGBMS:2005} and new results on the nonlinear state space models based on the GP~\citep{SvenssonS:2017,FrigolaLSR:2013,FrigolaCR:2014}. We also mention the nice overview by~\citet{Kocijan:2016}.

When the basic GP model is used for regression, it results in a computational complexity that is too high to be practically useful, stochastic inputs cannot be used and it cannot be used in an online fashion. These three fundamental problems of basic GP regression have been addressed in many different ways, which we return to in Section~\ref{s:LimitationsOfGPR}. The nonlinear system identification problem typically requires us to solve these three problems simultaneously. This brings us to our two main contributions of this paper. (1) We derive an algorithm allowing us to---in an online fashion---include stochastic training points to one of the classic sparse GP models, the so-called FITC (Fully Independent Training Conditional) algorithm. (2) We adapt the new algorithm to the nonlinear system identification problem, resulting in an online algorithm for nonlinear system identification. The experimental results show that the our new algorithm is indeed competitive compared to existing solutions. 

The system identification formulation takes inspiration from the nonlinear autoregressive model with exogenous (ARX) inputs of the following form
\begin{equation}
  \label{eq:SystemFunction}
  \ve{y_k} = \ve{\phi}(\ve{y_{k-1}}, \ldots, \ve{y_{k-n_y}}, \ve{u_{k-1}}, \ldots, \ve{u_{k-n_u}}),
\end{equation}
where $\ve{\phi}(\cdot)$ denotes some nonlinear function of past inputs $\ve{u_{k-1}}, \ldots, \ve{u_{k-n_u}}$ and past outputs $\ve{y_{k-1}}, \ldots, \ve{y_{k-n_y}}$ to the system. To this end we develop a non-parametric and probabilistic GP model which takes the following vector
\begin{align}
  \label{eq:ARXState}
  \ve{x_k} = (\ve{y_{k-1}}, \ldots, \ve{y_{k-n_y}}, \ve{u_{k-1}}, \ldots, \ve{u_{k-n_u}}),
\end{align}
as its input vector. The crucial part behind our solution from a system identification point of view is that we continuously keep track of the covariances between respective output estimates $\ve{y_k}, \ldots, \ve{y_{k-n_y}}$ and inputs $\ve{u_k}, \ldots, \ve{u_{k-n_u}}$. Every time we incorporate new training data, the respective means and covariances of these parameters are further refined.

The paper is organized as follows. Section~\ref{s:LimitationsOfGPR} starts by examining three important existing problems within GP regression, giving a quick summary of the solutions discussed in literature. In Section~\ref{s:StochasticMeasurements} we expand on these methods, enabling GP regression to be applied in an online manner using noisy input points. This results in the basic version of the new algorithm. Section~\ref{s:SonigDetails} subsequently outlines various ways of extending this algorithm, allowing it to be applied to system identification problems. Experimental results, first for the basic algorithm (Algorithm~\ref{alg:SONIG}) and then for its system identification set-up (Algorithm~\ref{alg:SISONIG}) are shown in Section~\ref{s:TestResults}. Section~\ref{s:Conclusions} finally gives conclusions and recommendations for future work along this direction.

\section{Backgrounds and limitations of Gaussian process regression}\label{s:LimitationsOfGPR}

Gaussian process regression is a powerful regression method but, like any method, it has its limitations. In this section we look at its background, what exact limitations it has and what methods are available in literature to tackle these. Also the assumptions made and the notation used is introduced.

\subsection{Regular Gaussian process regression} \label{ss:RegularGP}

GP regression~\citep{GPBook} is about approximating a function $f(\ve{x})$ through a number of $n$ training points (measurements) $(\ve{x_1},y_1), (\ve{x_2},y_2), \ldots, (\ve{x_n},y_n)$. Here, $\ve{x}$ denotes the training input point and $y$ the measured function output. (For now we assume scalar outputs. Section~\ref{ss:MultipleOutputs} looks at the multi-output case.) We assume that the training outputs are corrupted by noise, such that $y_i = f(\ve{x_i}) + \varepsilon$, with $\varepsilon \sim \N(0, \sigma_n^2)$ being Gaussian white noise with zero mean and variance~$\sigma_n^2$.

As shorthand notation, we merge all the training points $\ve{x_i}$ into a training set $X$ and all corresponding output values $y_i$ into an output vector $\ve{y}$. We now write the noiseless function output $f(X)$ as $\ve{f}$, such that $\ve{y} = \ve{f} + \ve{\varepsilon}$, with $\ve{\varepsilon} \sim \N(\ve{0}, \Sigma_n)$ and $\Sigma_n = \sigma_n^2 I$.

Once we have the training data, we want to predict the function value $f(\ve{x_*})$ at a specific test point $\ve{x_*}$. Equivalently, we can also predict the function values $\ve{f_*} = f(X_*)$ at a whole set of test points $X_*$. To accomplish this using GP regression, we assume that $\ve{f_*}$ and $\ve{f}$ have a prior joint Gaussian distribution given by
\begin{equation}
\begin{bmatrix}
\ve{f^0} \\
\ve{f_*^0}
\end{bmatrix} \sim \N\left(\begin{bmatrix}
m(X) \\
m(X_*)
\end{bmatrix}, \begin{bmatrix}
k(X,X) & k(X,X_*) \\
k(X_*,X) & k(X_*,X_*)
\end{bmatrix}\right) = \N\left(\begin{bmatrix}
\ve{m} \\
\ve{m_*}
\end{bmatrix}, \begin{bmatrix}
K_{} & K_{*} \\
K_{*}^T & K_{**}
\end{bmatrix}\right),\label{eq:GPStartingEquation}
\end{equation}
where in the second part of the equation we have introduced another shorthand notation. Note here that $m(\ve{x})$ is the prior mean function for the Gaussian process and $k(\ve{x},\ve{x'})$ is the prior covariance function. The superscript $0$ in $\ve{f^0}$ and $\ve{f_*^0}$ also denotes that we are referring to the prior distribution: no training points have been taken into account yet. In this paper we make no assumptions on the prior mean/kernel functions, but our examples will apply a zero mean function $m(\ve{x}) = 0$ and a squared exponential covariance function
\begin{equation}\label{eq:SECovarianceFunction}
k(\ve{x},\ve{x'}) = \alpha^2 \exp\left(-\frac{1}{2} \left(\ve{x} - \ve{x'}\right)^T \Lambda^{-1} \left(\ve{x} - \ve{x'}\right)\right),
\end{equation}
with $\alpha$ a characteristic output length scale and $\Lambda$ a diagonal matrix of characteristic input squared length scales. For now we assume that these hyperparameters are known, but in Section~\ref{ss:HyperparameterTuning} we look at ways to tune them.

From~\eqref{eq:GPStartingEquation} we can find the posterior distribution of both $\ve{f}$ and $\ve{f_*}$ given $\ve{y}$ as
\begin{align}\label{eq:RegularGPRegression}
\begin{bmatrix}
\ve{f^n} \\
\ve{f_*^n}
\end{bmatrix} & \sim \N\left(\begin{bmatrix}
\ve{\mu^n} \\
\ve{\mu_*^n}
\end{bmatrix}, \begin{bmatrix}
\Sigma_{}^n & \Sigma_{*}^n \\
\left(\Sigma_{*}^n\right)^T & \Sigma_{**}^n
\end{bmatrix}\right), \nonumber \\
\begin{bmatrix}
\ve{\mu^n} \\
\ve{\mu_*^n}
\end{bmatrix} & = \begin{bmatrix}
\left(K_{}^{-1} + \Sigma_n^{-1}\right)^{-1} \left(K_{}^{-1} \ve{m} + \Sigma_n^{-1} \ve{y}\right) \\
\ve{m_*} + K_{*}^T \left(K_{} + \Sigma_n\right)^{-1} \left(\ve{y} - \ve{m}\right)
\end{bmatrix}, \nonumber \\
\begin{bmatrix}
\Sigma_{}^n & \Sigma_{*}^n \\
\left(\Sigma_{*}^n\right)^T & \Sigma_{**}^n
\end{bmatrix} & = \begin{bmatrix}
\left(K_{}^{-1} + \Sigma_n^{-1}\right)^{-1} & \Sigma_n \left(K_{} + \Sigma_n\right)^{-1} K_{*} \\
K_{*}^T \left(K_{} + \Sigma_n\right)^{-1} \Sigma_n & K_{**} - K_{*}^T \left(K_{} + \Sigma_n\right)^{-1} K_{*}
\end{bmatrix}.
\end{align}
Note here that, while we use $\ve{m}$ and $K$ to denote properties of prior distributions, we use $\ve{\mu}$ and $\Sigma$ for posterior distributions. The superscript $n$ indicates these are posteriors taking $n$ training points into account, and while a star $*$ subscript denotes a parameter of the test set, an omitted subscript denotes a training parameter.

\subsection{Sparse Gaussian process regression} \label{ss:SparseGP}

An important limitation of Gaussian process regression is its computational complexity of $\ord(n^3)$, with $n$ the number of training points. This can be tackled through parallel computing~\citep{GalWR:2014,DeisenrothN:2015} but a more common solution is to use the so-called sparse methods. An overview of these is given by~\cite{SparseGPUnifyingView}, summarizing various contributions~\citep{SparseGreedyGP,SparseOnlineGP,FFSForSparseGP,SparseGPPseudoInputs} into a comprehensive framework. With these methods, and particularly with the FITC method that we will apply in this paper, the runtime can be reduced to being linear with respect to $n$, with only a limited reduction in how well the available data is being used.

All these sparse methods make use of so-called inducing input points $X_u$ to reduce the computational complexity. Such inducing input points are also used in the more recent work on variational inference~\citep{Titsias:2009,LatentVariableGP}. However, as pointed out by~\cite{McHutchonThesis}, these points are now not used for the sake of computational speed but merely as `information storage'. \cite{McHutchonThesis} also noted that the corresponding methods have a large number of parameters to optimize, making the computation of the derivatives rather slow. Furthermore, even though variations have been developed which do allow the application of variational inference to larger data sets~\citep{GPForBigData,GalWR:2014,GPVariationalInference}, online methods generally require simpler and faster methods, like the NIGP (Noisy Input GP) method from~\cite{GPWithInputNoise}, which is what we will focus on.

To apply sparse GP regression, we first find the posterior distribution of the inducing outputs $\ve{f_u}$ at the corresponding inducing input points $X_u$. This can be done in $\ord(n^3)$ time through~\eqref{eq:RegularGPRegression} (replacing $\ve{f_*}$ by $\ve{f_u}$) or in $\ord(nn_u^2)$ time through the FITC regression equation
\begin{align}\label{eq:FITCEquations}
\begin{bmatrix}
\ve{f^n} \\
\ve{f_u^n}
\end{bmatrix} & \sim \N\left(\begin{bmatrix}
\ve{\mu^n} \\
\ve{\mu_u^n}
\end{bmatrix}, \begin{bmatrix}
\Sigma_{}^n & \Sigma_{u}^n \\
\left(\Sigma_{u}^n\right)^T & \Sigma_{uu}^n
\end{bmatrix}\right), \nonumber \\
\begin{bmatrix}
\ve{\mu^n} \\
\ve{\mu_u^n}
\end{bmatrix} & = \begin{bmatrix}
\ve{m} + \Sigma_{}^n \Sigma_n^{-1} \left(\ve{y} - \ve{m}\right) \\
\ve{m_u} + \left(\Sigma_{u}^n\right)^T \left(\Lambda_n^{-1} + \Sigma_n^{-1}\right) \left(\ve{y} - \ve{m}\right)
\end{bmatrix}, \nonumber \\
\Sigma_{}^n & = \left(\Lambda_n^{-1} + \Sigma_n^{-1}\right)^{-1} + \Sigma_n \left(\Lambda_n + \Sigma_n\right)^{-1} K_{u} \Delta^{-1} K_{u}^T \left(\Lambda_n + \Sigma_n\right)^{-1} \Sigma_n, \nonumber \\
\Sigma_{u}^n & = \Sigma_n \left(\Lambda_n + \Sigma_n\right)^{-1} K_{u} \Delta^{-1} K_{uu}, \nonumber \\
\Sigma_{uu}^n & = K_{uu} \Delta^{-1} K_{uu}.
\end{align}
Here we have used the shorthand notation $\Delta = K_{uu} + K_{u}^T \left(\Lambda_n + \Sigma_n\right)^{-1} K_{u}$ and $\Lambda_n = \diag\left(K_{} - K_{u}^T K_{uu}^{-1} K_{u}\right)$, with $\diag$ being the function that sets all non-diagonal elements of the given matrix to zero. The FITC regression equation is an approximation, based on the assumption that all function values $f(\ve{x_1}), \ldots, f(\ve{x_n})$ are (a priori) independent given $\ve{f_u}$.

Once we know the distribution of $\ve{f_u}$, we calculate the posterior distribution of $\ve{f_*}$. Mathematically, this method is equivalent to assuming that $\ve{f}$ and $\ve{f_*}$ are conditionally independent, given $\ve{f_u}$. It follows that
\begin{align}\label{eq:SparseGPStep2}
\begin{bmatrix}
\ve{f_u^n} \\
\ve{f_*^n}
\end{bmatrix} & \sim \N\left(\begin{bmatrix}
\ve{\mu_u^n} \\
\ve{\mu_*^n}
\end{bmatrix}, \begin{bmatrix}
\Sigma_{uu}^n & \Sigma_{u*}^n \\
\Sigma_{*u}^n & \Sigma_{**}^n
\end{bmatrix}\right), \nonumber \\
\begin{bmatrix}
\ve{\mu_u^n} \\
\ve{\mu_*^n}
\end{bmatrix} & = \begin{bmatrix}
\ve{\mu_u^n} \\
\ve{m_*} + K_{*u} K_{uu}^{-1} \left(\ve{\mu_u^n} - \ve{m_u}\right)
\end{bmatrix}, \nonumber \\
\begin{bmatrix}
\Sigma_{uu}^n & \Sigma_{u*}^n \\
\Sigma_{*u}^n & \Sigma_{**}^n
\end{bmatrix} & = \begin{bmatrix}
\Sigma_{uu}^n & \Sigma_{uu}^n K_{uu}^{-1} K_{u*} \\
K_{*u} K_{uu}^{-1} \Sigma_{uu}^n & K_{**} - K_{*u} K_{uu}^{-1} \left(K_{uu} - \Sigma_{uu}^n\right) K_{uu}^{-1} K_{u*}
\end{bmatrix}.
\end{align}

\subsection{Online Gaussian process regression}

A second limitation of GP regression is the difficulty with which it can incorporate new training points. For regular GP regression, a new measurement $n+1$ can be added to the existing set of $n$ training points through a matrix update, resulting in an $\ord(n^2)$ runtime. For sparse methods using inducing input (basis) points this can generally be done more efficiently \citep{SparseOnlineGP, SparseGPUnifyingView, OnlineSparseGPInImageProcessing, SparseOnlineGPForWind, GPForBigData}. The main downside is that most methods set requirements on these inducing input points. However, the FITC and the PITC (Partially Independent Training Conditional) methods can be set up in an online way without such constraints \citep{RecursiveGPConferencePaper, RecursiveGaussianProcess,OnlineSparseGP}. We briefly summarize the resulting algorithm.

Suppose that we know the distribution of $\ve{f_u}$, given the first $n$ training points. This is written as $\ve{f_u^n}$. Next, consider a new measurement $(\ve{x_{n+1}}, y_{n+1})$, whose notation we will shorten to $(\ve{x_+}, y_+)$. To incorporate it, we first predict the posterior distribution of $f_+ = f(\ve{x_+})$ based on only the first $n$ training points. Identically to~\eqref{eq:SparseGPStep2}, this results in
\begin{align}\label{eq:SparseOnlineGPPrediction}
\begin{bmatrix}
\ve{f_u^n} \\
f_+^n
\end{bmatrix} & \sim \N\left(\begin{bmatrix}
\ve{\mu_u^n} \\
\mu_+^n
\end{bmatrix}, \begin{bmatrix}
\Sigma_{uu}^n & \Sigma_{u+}^n \\
\Sigma_{+u}^n & \Sigma_{++}^n
\end{bmatrix}\right), \nonumber \\
\begin{bmatrix}
\ve{\mu_u^n} \\
\mu_+^n
\end{bmatrix} & = \begin{bmatrix}
\ve{\mu_u^n} \\
m_+ + K_{+u} K_{uu}^{-1} \left(\ve{\mu_u^n} - \ve{m_u}\right)
\end{bmatrix}, \nonumber \\
\begin{bmatrix}
\Sigma_{uu}^n & \Sigma_{u+}^n \\
\Sigma_{+u}^n & \Sigma_{++}^n
\end{bmatrix} & = \begin{bmatrix}
\Sigma_{uu}^n & \Sigma_{uu}^n K_{uu}^{-1} K_{u+} \\
K_{+u} K_{uu}^{-1} \Sigma_{uu}^n & K_{++} - K_{+u} K_{uu}^{-1} \left(K_{uu} - \Sigma_{uu}^n\right) K_{uu}^{-1} K_{u+}
\end{bmatrix}.
\end{align}
If we subsequently incorporate the new measurement, identically to~\eqref{eq:RegularGPRegression}, then we get
\begin{align}\label{eq:SparseOnlineGPUpdate}
\begin{bmatrix}
\ve{f_u^{n+1}} \\
f_+^{n+1}
\end{bmatrix} & \sim \N\left(\begin{bmatrix}
\ve{\mu_u^{n+1}} \\
\mu_+^{n+1}
\end{bmatrix}, \begin{bmatrix}
\Sigma_{uu}^{n+1} & \Sigma_{u+}^{n+1} \\
\Sigma_{+u}^{n+1} & \Sigma_{++}^{n+1}
\end{bmatrix}\right), \nonumber \\
\begin{bmatrix}
\ve{\mu_u^{n+1}} \\
\mu_+^{n+1}
\end{bmatrix} & = \begin{bmatrix}
\ve{\mu_u^n} + \Sigma_{u+}^n \left(\Sigma_{++}^n + \sigma_n^2\right)^{-1} \left(y_+ - \mu_+^n\right) \\
\sigma_n^2 \left(\Sigma_{++}^n + \sigma_n^2\right)^{-1} \mu_+^n + \Sigma_{++}^n \left(\Sigma_{++}^n + \sigma_n^2\right)^{-1} y_+
\end{bmatrix}, \nonumber \\
\begin{bmatrix}
\Sigma_{uu}^{n+1} & \Sigma_{u+}^{n+1} \\
\Sigma_{+u}^{n+1} & \Sigma_{++}^{n+1}
\end{bmatrix} & = \begin{bmatrix}
\Sigma_{uu}^n - \Sigma_{u+}^n (\Sigma_{++}^n + \sigma_n^2)^{-1} \Sigma_{+u}^n & \Sigma_{u+}^n (\Sigma_{++}^n + \sigma_n^2)^{-1} \sigma_n^2 \\
\sigma_n^2 (\Sigma_{++}^n + \sigma_n^2)^{-1} \Sigma_{+u}^n & \sigma_n^2 (\Sigma_{++}^n + \sigma_n^2)^{-1} \Sigma_{++}^n
\end{bmatrix}.
\end{align}
This expression (or at least the part relating to $\ve{f_u}$) is the update law for the FITC algorithm. It tells us exactly how the distribution of $\ve{f_u^{n+1}}$ (both $\ve{\mu_u^{n+1}}$ and $\Sigma_{uu}^{n+1}$) depends on $\ve{x_+}$ and $y_+$. With this distribution, we can subsequently always find the distribution of new test point outputs $\ve{f_*}$ in an efficient way through~\eqref{eq:SparseGPStep2}.

\subsection{Using stochastic input points}

The third limitation is that the GP regression algorithm assumes that the input points are deterministic. This assumption concerns both the training (measurement) points $\ve{x}$ and the test points $\ve{x_*}$. For noisy (stochastic) test points $\ve{x_*}$, we can work around the problem by applying moment matching~\citep{DeisenrothThesis}. This technique can subsequently also be expanded for noisy training points~\citep{GPUncertainPoints}, but the effectiveness is limited because the method integrates over all possible \emph{a priori} functions, and not over all possible \emph{a posteriori} functions. There are methods that include posterior distributions~\citep{GPTaylorApproximation} but these often only work for noisy test points and not for noisy training points. The NIGP  algorithm~\citep{GPWithInputNoise} is the only previously existing algorithm that we are aware of that is include posterior distributions at the same time as it can deal with noisy training points. This is the method we will be expanding upon in this paper.

It should be noted that the variational methods mentioned earlier can also deal with stochastic input points, up to a certain degree. However, as also mentioned before, they cannot do so as computationally efficient as the NIGP algorithm or the algorithm that we will develop, so their applicability to online system identification remains limited. Additionally, it is also possible to take into account the effects of noisy training points by assuming that the noise variance varies over the input space; a feature called heteroscedasticity. This has been investigated quite in-depth \citep{GPInputDependentNoise,HeteroscedasticGP,HeteroscedasticGPWithNonGaussianResiduals,GPVariableNoise} but it would give more degrees of freedom to the learning algorithm than would be required, and as a result these methods have a reduced performance for the problems we consider. We will not consider these methods further in this paper.

\section{Expanding the algorithm for stochastic training points}\label{s:StochasticMeasurements}

This section contains our main contribution: enabling the FITC algorithm to handle stochastic training points in an online way. From a computational point of view, the novel update laws given here are simple and efficient, relative to other methods.

\subsection{The online stochastic measurement problem}

Consider the case where we know the distribution $\ve{f_u^n} \sim \N\left(\ve{\mu_u^n, \Sigma_{uu}^n}\right)$ (initially we have $\ve{f_u^0} \sim \N\left(\ve{\mu_u^0}, \Sigma_{uu}^0\right) = \N\left(\ve{m_u}, K_{uu}\right)$) and we obtain a new measurement at some unknown input point $\ve{x_+}$. As before, the true function output $f_+ = f(\ve{x_+})$ is also unknown. Our measurement only gives us values $\ve{\hat{x}_+}$ and $\hat{f}_+$ approximating these and hence tells us that $\ve{x_+} \sim \N(\ve{\hat{x}_+}, \Sigma_{+_x})$ and $f_+ \sim \N(\hat{f}_+, \Sigma_{+_f})$. (Note that $\hat{f}_+$ and $y_+$ are identical, and so are $\Sigma_{+_f}$ and $\sigma_n^2$. For the sake of uniform notation, we have renamed them.) We assume that the noise on the input and the output is independent, and hence that $\ve{x_+}$ and $f_+$ are a priori not correlated.

Our main goal is to find the posterior distribution $\ve{f_u^{n+1}}$, given this stochastic training point. This can be done through
\begin{equation}\label{eq:PosteriorDistribution}
p(\ve{f_u^{n+1}}|\ve{\hat{x}_+},\hat{f}_+,\ve{f_u^n}) = \int_X p(\ve{f_u^{n+1}}|\ve{x_+},\ve{\hat{x}_+},\hat{f}_+,\ve{f_u^n}) p(\ve{x_+}|\ve{\hat{x}_+},\hat{f}_+,\ve{f_u^n}) \, d\ve{x_+}.
\end{equation}
In the integral, the first probability $p(\ve{f_u^{n+1}}|\ve{x_+},\ve{\hat{x}_+},\hat{f}_+,\ve{f_u^n})$ is the update law for $\ve{f_u^{n+1}}$ if $\ve{x_+}$ was known exactly. It directly follows from~\eqref{eq:SparseOnlineGPUpdate}. The second probability $p(\ve{x_+}|\ve{\hat{x}_+},\hat{f}_+,\ve{f_u^n})$ is the posterior distribution of $\ve{x_+}$, given both $\ve{f_u^n}$ and the new measurement. Since this latter term is more difficult to deal with, we examine it first.

\subsection{The posterior distribution of the training point}\label{ss:PosteriorMeasurementDistribution}

The posterior distribution of $\ve{x_+}$ can be found through Bayes' theorem,
\begin{equation}
p(\ve{x_+}|\hat{f}_+,\ve{\hat{x}_+},\ve{f_u^n}) = \frac{p(\hat{f}_+|\ve{x_+},\ve{\hat{x}_+},\ve{f_u^n}) p(\ve{x_+}|\ve{\hat{x}_+},\ve{f_u^n})}{p(\hat{f}_+|\ve{\hat{x}_+},\ve{f_u^n})}.
\end{equation}
Here $p(\ve{x_+}|\ve{\hat{x}_+},\ve{f_u^n}) = p(\ve{x_+}|\ve{\hat{x}_+}) = \N(\ve{\hat{x}_+}, \Sigma_{+_x})$ and $p(\hat{f}_+|\ve{\hat{x}_+},\ve{f_u^n})$ equals an unknown constant (i.e., not depending on $\ve{x_+}$). Additionally, 
\begin{equation}\label{eq:ConditionalMeasurementProbability}
p(\hat{f}_+|\ve{x_+},\ve{\hat{x}_+},\ve{f_u^n}) = p(\hat{f}_+|\ve{x_+},\ve{f_u^n}) = \N\left(\mu_+^n, \Sigma_{++}^n + \Sigma_{+_f}\right),
\end{equation}
where $\mu_+^n$ and $\Sigma_{++}^n$ follow from~\eqref{eq:SparseOnlineGPPrediction}. Note that both these quantities depend on $\ve{x_+}$ in a nonlinear way. Because of this, the resulting probability $p(\ve{x_+}|\hat{f}_+,\ve{\hat{x}_+},\ve{f_u^n})$ will be non-Gaussian. To work around this problem, we have to make some simplifying assumptions. Similarly to~\cite{GPTaylorApproximation}, we linearize the Gaussian process $\hat{f}_+$ (which depends on $\ve{x_+}$) around a point $\ve{\bar{x}_+}$. That is, we assume that
\begin{equation}\label{eq:LinearizedGPAssumption}
p(\hat{f}_+|\ve{x_+},\ve{f_u^n}) = \N\left(\mu_+^n(\ve{\bar{x}_+}) + \frac{d\mu_+^n(\ve{\bar{x}_+})}{d\ve{x_+}} \left(\ve{x_+} - \ve{\bar{x}_+}\right), \Sigma_{++}^n(\ve{\bar{x}_+}) + \Sigma_{+_f}\right).
\end{equation}
In other words, we assume that the mean varies linearly with $\ve{x_+}$, while the covariance is constant everywhere. The solution for $p(\ve{x_+}|\hat{f}_+,\ve{\hat{x}_+},\ve{f_u^n})$ is now Gaussian and follows as
\begin{align}\label{eq:PosteriorMeasurementPointDistribution}
p(\ve{x_+}|\hat{f}_+,\ve{\hat{x}_+},\ve{f_u^n}) & = \N\left(\ve{\hat{x}_+^{n+1}},\Sigma_{+_x}^{n+1}\right), \nonumber \\
\ve{\hat{x}_+^{n+1}} & = \ve{\hat{x}_+} + \Sigma_{+_x}^{n+1} \left(\left(\frac{d\mu_+^n(\ve{\bar{x}_+})}{d\ve{x_+}}\right)^T \left(\Sigma_{++}^n(\ve{\bar{x}_+}) + \Sigma_{+_f}\right)^{-1} \right. \nonumber \\
& \hspace{80pt} \left.\left(\frac{d\mu_+^n(\ve{\bar{x}_+})}{d\ve{x_+}} \left(\ve{\bar{x}_+} - \ve{\hat{x}_+}\right) + \left(\hat{f}_+ - \mu_+^n(\ve{\bar{x}_+})\right)\right)\right), \nonumber \\
\Sigma_+^{n+1} & = \left(\left(\frac{d\mu_+^n(\ve{\bar{x}_+})}{d\ve{x_+}}\right)^T \left(\Sigma_{++}^n(\ve{\bar{x}_+}) + \Sigma_{+_f}\right)^{-1} \left(\frac{d\mu_+^n(\ve{\bar{x}_+})}{d\ve{x_+}}\right) + \Sigma_{+_x}^{-1}\right)^{-1}.
\end{align}
We are left to choose a linearization point $\ve{\bar{x}_+}$. The above equation is easiest to apply when we choose $\ve{\bar{x}_+} = \ve{\hat{x}_+}$, but when $\left(\hat{f}_+ - \mu_+^n(\ve{\bar{x}_+})\right)$ is large, this may result in inaccuracies due to the linearization. It is generally more accurate to choose $\ve{\bar{x}_+}$ as $\ve{\hat{x}_+^{n+1}}$. However, $\ve{\hat{x}_+^{n+1}}$ is initially unknown, so it may be necessary to apply the above equation multiple times, each time resetting $\ve{\bar{x}_+}$ to the latest value of $\ve{\hat{x}_+^{n+1}}$ that was found, to find the most accurate posterior distribution of $\ve{x_+}$.

\subsection{Updating the inducing input point function values}\label{ss:SONIGUpdateLaw}

Using the above, we can solve~\eqref{eq:PosteriorDistribution}. This is done by approximating the GP $\ve{f_u}(\ve{x_+})$ by its Taylor expansion around $\ve{\hat{x}_+^{n+1}}$. Element-wise we write this as
\begin{align}\label{eq:TaylorExpansion}
f_{u_i}^{n+1}(\ve{x_+}) & = f_{u_i}^{n+1}(\ve{\hat{x}_+^{n+1}}) + \frac{df_{u_i}^{n+1}(\ve{\hat{x}_+^{n+1}})}{d\ve{x_+}} \left(\ve{x_+} - \ve{\hat{x}_+^{n+1}}\right) \nonumber \\
& \hspace{40pt} + \frac{1}{2} \left(\ve{x_+} - \ve{\hat{x}_+^{n+1}}\right)^T \frac{d^2f_{u_i}^{n+1}(\ve{\hat{x}_+^{n+1}})}{d\ve{x}_+^2} \left(\ve{x_+} - \ve{\hat{x}_+^{n+1}}\right) + \ldots.
\end{align}
Within this Taylor expansion,~\cite{GPTaylorApproximation} made the assumption that higher order derivatives like $\frac{d^2 \ve{f_u^n}}{d\ve{x}_+^2}$ are negligible, remaining with just a linearization of the GP. We do not make this assumption, but instead assume that $\Sigma_{+_x}^2$ and higher powers of $\Sigma_{+_x}$ are negligible. (If the uncertainties in $\ve{x_+}$ are so large that this assumption does not hold, then any form of Gaussian process regression is likely to fail.) This assumption is not only more loose---resulting in an extra term in~\eqref{eq:TaylorExpansion}---but it is also easier to verify.

An additional assumption we need to make is that $\ve{x_+}$ is independent of $\ve{f_u}$. This is reasonable, as $\ve{x_+}$ is only contaminated by Gaussian white noise. Applying this, we can solve~\eqref{eq:PosteriorDistribution} through both~\eqref{eq:PosteriorMeasurementPointDistribution} and~\eqref{eq:TaylorExpansion}. The result equals
\begin{align}\label{eq:SONIGUpdate}
\ve{f_u^{n+1}} & \sim \N\left(\ve{\mu_u^{n+1}}, \Sigma_{uu}^{n+1}\right), \nonumber \\
\mu_{u_i}^{n+1} & = \mu_{u_i}^{n+1}(\ve{\hat{x}_+^{n+1}}) + \frac{1}{2} \tr\left(\frac{d^2 \mu_{u_i}^{n+1}(\ve{\hat{x}_+^{n+1}})}{d\ve{x}_+^2} \Sigma_{+_x}^{n+1}\right), \nonumber \\
\Sigma_{u_iu_j}^{n+1} & = \Sigma_{u_iu_j}^{n+1}(\ve{\hat{x}_+^{n+1}}) + \left(\frac{d\mu_{u_j}^{n+1}(\ve{\hat{x}_+^{n+1}})}{d\ve{x_+}}\right) \Sigma_{+_x}^{n+1} \left(\frac{d\mu_{u_i}^{n+1}(\ve{\hat{x}_+^{n+1}})}{d\ve{x_+}}\right)^T \nonumber \\
& \hspace{120pt} + \frac{1}{2} \tr\left(\left(\frac{d^2 \Sigma_{u_iu_j}^{n+1}(\ve{\hat{x}_+^{n+1}})}{d\ve{x}_+^2}\right) \Sigma_{+_x}^{n+1}\right).
\end{align}
Here, the functions $\mu_{u_i}^{n+1}(\ve{x_+})$ and $\Sigma_{u_iu_j}^{n+1}(\ve{x_+})$ (for a given point $\ve{x_+}$) are given by~\eqref{eq:SparseOnlineGPUpdate}, combined with~\eqref{eq:SparseOnlineGPPrediction}. Finding all the derivatives of these parameters can be a daunting task, especially for non-scalar inputs $\ve{x_+}$, but the mathematics are relatively straightforward, so for this we refer to the Appendix.

It is interesting to compare expression~\eqref{eq:SONIGUpdate} with what was used by~\cite{GPWithInputNoise} in their NIGP algorithm. They did not include the term involving $d^2\mu_{u_i}^{n+1}/d\ve{x}_+^2$. Later on, in Section~\ref{ss:SonigTest}, we will find that exactly this term causes the new algorithm to perform better than the NIGP algorithm. As such, the above update law~\eqref{eq:SONIGUpdate} also serves as an improvement with respect to the NIGP algorithm.

\subsection{The SONIG algorithm} \label{ss:SonigAlgorithm}

Applying the equations developed so far is done through the Sparse Online Noisy Input GP (SONIG) algorithm, outlined in Algorithm~\ref{alg:SONIG}. This algorithm is computationally efficient, in the sense that a single updating step (incorporating one training point) can be done in constant runtime with respect to the number of training points already processed. The runtime does depend on the number of inducing input points through $\ord(n_u^3)$, just like it does for all sparse GP regression algorithms.

\begin{algorithm}[!htp]
	\label{alg:SONIG}
	\SetKwBlock{Input}{Input:}{}
	\Input{
		A possibly expanding set of training points $(\ve{x_1},y_1), \ldots, (\ve{x_n},y_n)$ in which both $\ve{x}$ and $y$ are distorted by Gaussian white noise.
	}
	\SetKwBlock{Prepare}{Preparation:}{}
	\Prepare{
		Either choose the hyperparameters based on expert knowledge, or apply the NIGP hypertuning methods of~\cite{GPWithInputNoise} on a subset of the data (a few hundred points) to find the hyperparameters. \\
		Optionally, apply the NIGP regression methods on this subset of data to obtain an initial distribution of $\ve{f_u}$. Otherwise, initialize $\ve{f_u}$ as $\N(\ve{m_u},K_{uu})$.
	}
	\SetKwBlock{Update}{Updating:}{}
	\Update{
		\While{there are unprocessed training points $(\ve{x_{n+1}}, y_{n+1})$}{
		1. Apply~\eqref{eq:PosteriorMeasurementPointDistribution} to find the posterior distribution of the training point $\ve{x_{n+1}}$ (written as $\ve{x_+}$).\\ 2. Use~\eqref{eq:SONIGUpdate} to update the distribution of $\ve{f_u}$.\\
		3. Optionally, use~\eqref{eq:PosteriorOutputDistribution} and~\eqref{eq:PosteriorOutputCovariance} to calculate the posterior distribution of the function value $\ve{f}(\ve{x_+})$ (written as $\ve{f_+}$).}
	}
	\SetKwBlock{Predict}{Prediction:}{}
	\Predict{
		 Apply~\eqref{eq:SparseGPStep2} to find the distribution $\ve{f_*}$ for any set of deterministic test points. For stochastic test points, use the expansion from Section~\ref{ss:StochasticTestPointPredictions}.
	}
	\hspace{1pt}\\ 
	\hspace{1pt}\\ 
	\caption{The Sparse Online Noisy Input GP (SONIG) algorithm: an online version of the FITC algorithm capable of dealing with stochastic (noisy) training points.}
\end{algorithm}

\section{Extensions of the SONIG algorithm}\label{s:SonigDetails}

In the previous section we have presented the basic idea behind the SONIG algorithm. There are various further extensions that can be derived and implemented in the algorithm. For instance, the algorithm can deal with multi-output functions $\ve{f}(\ve{x})$ (Section~\ref{ss:MultipleOutputs}), it can give us the posterior distribution of the output $\ve{f_+}$ as well as its correlation with the input $\ve{x_+}$ (Section~\ref{ss:PosteriorInputOutputCovariance}), we can implement hyperparameter tuning (Section~\ref{ss:HyperparameterTuning}), we can add inducing input points online (Section~\ref{ss:AddingInducingInputPoints}) and we can make predictions $\ve{f_*}$ using stochastic test points $\ve{x_*}$ (Section~\ref{ss:StochasticTestPointPredictions}). Many of these extensions are necessary to apply the SONIG algorithm for system identification. The resulting system identification algorithm is summarized in Algorithm~\ref{alg:SISONIG}.

\subsection{Multiple outputs}\label{ss:MultipleOutputs}

So far we have approximated functions $f(\ve{x})$ with only one output. It is also possible to approximate functions $\ve{f}(\ve{x})$ with $d_y > 1$ outputs. A common way in which this is done in GP regression algorithms~\citep{PILCO,MultiOutputGP} is by assuming that, given a deterministic input $\ve{x}$, all outputs $f_1(\ve{x}), \ldots, f_{d_y}(\ve{x})$ are independent. With this assumption, it is possible to keep a separate inducing input point distribution $\ve{f_u^i} \sim \N\left(\ve{\mu_u^i}, \Sigma_u^i\right)$ for each output $f_i(\ve{x})$. Hence, each output is basically treated separately.

When using stochastic input points (again, see~\cite{PILCO}) the outputs do become correlated. We now have two options. If we take this correlation into account, we have to keep track of the joint distribution of the vectors $\ve{f_u^1}, \ldots, \ve{f_u^{d_y}}$, effectively merging them into one big vector. This results in a vector of size $n_ud_y$, giving our algorithm a computational complexity of $\ord(n_u^3d_y^3)$. Alternatively, we could also neglect the correlation between the inducing input point distributions $\ve{f_u^i}$ caused by stochastic training points $\ve{x_+} \sim \N(\ve{\hat{x}_+}, \Sigma_{+_x})$. If we do, we can continue to treat each function output separately, giving our algorithm a runtime of $\ord(n_u^3d_y)$. Because one of our aims in this paper is to reduce the runtime of GP regression algorithms, we will apply the second option.

When each output is treated separately, each output also has its own hyperparameters. Naturally, the prior output covariance $\alpha_i^2$ and the output noise $\sigma_{n_i}^2$ can differ per output $f^i$, but also the input length scales $\Lambda_i$ may be chosen differently for each output. In fact, it is even possible to specify a fully separate covariance function $k_i(\ve{x},\ve{x'})$ per output $f^i$, though in this paper we stick with the squared exponential covariance function.

Naturally, there are a few equations which we should adjust slightly in the case of multivariate outputs. In particular, in equation~\eqref{eq:PosteriorMeasurementPointDistribution}, the parameter $\Sigma_{++}^n(\ve{\bar{x}_+})$ would not be a scalar anymore, but become a matrix. Due to our assumption that the outputs are independent, it would be a diagonal matrix. Similarly, the derivative $d\mu_+^n/d\ve{x_+}$ would not be a row vector anymore. Instead, it would turn into the matrix $d\ve{\mu_+^n}/d\ve{x_+}$. With these adjustments,~\eqref{eq:PosteriorMeasurementPointDistribution} still holds and all other equations can be applied as usual.

\subsection{The posterior distribution of the measured output}\label{ss:PosteriorInputOutputCovariance}

In Section~\ref{ss:SONIGUpdateLaw} we found the posterior distribution for $\ve{f_u}$. For some applications (like the system identification set-up presented in Section~\ref{ss:MethodApplication}) we also need the posterior distribution of the measured function value $\ve{f_+}$, even though we do not exactly know to which input it corresponds. We can find this element-wise, using the same methods, through
\begin{align}\label{eq:PosteriorOutputDistribution}
\ex[\ve{f_+}]_i & = \mu_{+_i}^{n+1}(\ve{\hat{x}_+^{n+1}}) + \frac{1}{2} \tr\left(\frac{d^2 \mu_{+_i}^{n+1}(\ve{\hat{x}_+^{n+1}})}{d\ve{x}_+^2} \Sigma_{+_x}^{n+1}\right), \nonumber \\
\va[\ve{f_+},\ve{f_+}]_{i,j} & = \Sigma_{+_i+_j}^{n+1}(\ve{\hat{x}_+^{n+1}}) + \left(\frac{d\mu_{+_i}^{n+1}(\ve{\hat{x}_+^{n+1}})}{d\ve{x_+}}\right) \Sigma_{+_x}^{n+1} \left(\frac{d\mu_{+_j}^{n+1}(\ve{\hat{x}_+^{n+1}})}{d\ve{x_+}}\right)^T \nonumber \\
& \hspace{120pt} + \frac{1}{2} \tr\left(\left(\frac{d^2 \Sigma_{+_i+_j}^{n+1}(\ve{\hat{x}_+^{n+1}})}{d\ve{x}_+^2}\right) \Sigma_{+_x}^{n+1}\right).
\end{align}
Note here that $\Sigma_{++}^{n+1}(\ve{\hat{x}_+^{n+1}})$ is (by assumption) a diagonal matrix, simplifying the above equation for non-diagonal terms. As such, the covariance between two different function outputs $f_{+_i}$ and $f_{+_j}$ only depends on the second term in the above expression. 

It may occur that we also need to know the posterior covariance between the function value $\ve{f_+}$ and the function input $\ve{x_+}$. Using the same method, we can find that
\begin{equation}\label{eq:PosteriorOutputCovariance}
\va[\ve{f_+},\ve{x_+}] = \left(\frac{d \ve{\mu_+^{n+1}}(\ve{\hat{x}_+^{n+1}})}{d\ve{x_+}}\right) \Sigma_{+_x}^{n+1}.
\end{equation}
This allows us to find the joint posterior distribution of $\ve{x_+}$ and $\ve{f_+}$.

\subsection{Applying hyperparameter tuning to the algorithm} \label{ss:HyperparameterTuning}

So far we have assumed that the hyperparameters of the Gaussian process are known a priori. When this is not the case, they need to be tuned first. While this could be done using expert knowledge of the system, it can also be done automatically.

\cite{GPWithInputNoise}, with their NIGP method, offer an effective method of tuning the hyperparameters, which also tells us the input noise variance $\Sigma_{+_x}$. However, this method has a computational complexity of $\ord(n^3)$, with $n$ still the number of training points. As such, it can only be used for a small number of measurements and it cannot be used online. Hence NIGP does not seem to be applicable to our problem.

However,~\cite{SparseGPComparison} compare various GP regression algorithms, including methods to tune hyperparameters. One of their main conclusions is that the subset-of-data (SoD) method provides a good trade-off between computational complexity and prediction accuracy. When using the SoD method, we do not apply hyperparameter tuning to our full data set, of possibly tens of thousands of input-output pairs. Instead, we randomly take a subset (say, a few hundred) of these data points and tune the hyperparameters only based on this selection. For this latter step any suitable hyperparameter tuning algorithm can be used, although we use the NIGP method described by~\cite{GPWithInputNoise}.

\cite{SparseGPComparison} also conclude that, for the regression problem with known hyperparameters, the FITC algorithm provides a very good trade-off between complexity and accuracy. So after having tuned the hyperparameters, it will be an effective choice to use the online FITC algorithm with stochastic training points (that is, the SONIG algorithm) as our regression method.

\subsection{Adjusting the set of inducing input points online}\label{ss:AddingInducingInputPoints}

When using an online GP regression algorithm, it is often not known in advance what kind of measured input points $\ve{x}$ the system will get. As such, choosing the inducing input points $X_u$ in advance is not always possible, nor wise. Instead, we can adjust the inducing input points while the algorithm is running. There are ways to fully tune the set of inducing input points, like using the latent variable methods by~\cite{Titsias:2009,LatentVariableGP}, but those methods require the optimization of many parameters, resulting in a computationally complex procedure. To keep the required computations limited, we have to opt for a simpler method and add/remove inducing input points based on areas of the input space we are interested in or have data at.

Suppose that we denote the current set of inducing input points by $X_u$, and that we want to add an extra set of inducing input points $X_{u_+}$. In this case, given all the data that we have, the distributions of $\ve{f_u}$ and $\ve{f_{u_+}}$ satisfy, identically to~\eqref{eq:SparseGPStep2},
\begin{align}\label{eq:AddingInducingInputPoints}
\begin{bmatrix}
\ve{f_u} \\
\ve{f_{u_+}}
\end{bmatrix} & \sim \N\left(\begin{bmatrix}
\ve{\mu_u} \\
\ve{\mu_{u_+}}
\end{bmatrix}, \begin{bmatrix}
\Sigma_{uu} & \Sigma_{uu_+} \\
\Sigma_{u_+u} & \Sigma_{u_+u_+}
\end{bmatrix}\right), \nonumber \\
\begin{bmatrix}
\ve{\mu_u} \\
\ve{\mu_{u_+}}
\end{bmatrix} & = \begin{bmatrix}
\ve{\mu_u} \\
\ve{m_{u_+}} + K_{u_+u} K_{uu}^{-1} \left(\ve{\mu_u} - \ve{m_u}\right)
\end{bmatrix}, \nonumber \\
\begin{bmatrix}
\Sigma_{uu}^n & \Sigma_{uu_+}^n \\
\Sigma_{u_+u}^n & \Sigma_{u_+u_+}^n
\end{bmatrix} & = \begin{bmatrix}
\Sigma_{uu}^n & \Sigma_{uu}^n K_{uu}^{-1} K_{uu_+} \\
K_{u_+u} K_{uu}^{-1} \Sigma_{uu}^n & K_{u_+u_+} - K_{u_+u} K_{uu}^{-1} \left(K_{uu} - \Sigma_{uu}^n\right) K_{uu}^{-1} K_{uu_+}
\end{bmatrix}.
\end{align}
With this combined set of old and new inducing input points, we can then continue incorporating new training points without losing any data.

Additionally, it is possible to remove unimportant inducing input points when desired. An inducing input point can be `unimportant' when it does not provide much information (it contributes little to the log-likelihood) or when it provides information we are not interested in, for instance when it lies in a part of the input space we do not care about. In this case, its entry can simply be removed from $\ve{f_u}$. Since it is possible to both add and remove inducing input points, it is naturally also possible to shift them around (first add new points, then remove old points) whenever deemed necessary.

The way in which we add inducing input points in the SONIG algorithm is as follows. Whenever we incorporate a new training point with posterior input distribution $\ve{x_+} \sim \N\left(\ve{\hat{x}_+^{n+1}}, \Sigma_{+_x}^{n+1}\right)$, we check if $\ve{\hat{x}_+^{n+1}}$ is already close to any existing inducing input point. To be more precise, we examine the normalized squared distance
\begin{equation}
\left(\ve{\hat{x}_+^{n+1}} - \ve{x_{u_i}}\right)^T \Lambda^{-1} \left(\ve{\hat{x}_+^{n+1}} - \ve{x_{u_i}}\right)
\end{equation}
for each inducing input point $\ve{x_{u_i}}$. If there is no inducing input point whose normalized squared distance is below a given threshold (often chosen to be roughly $1$, but tuned to get a satisfactory number of points), then it means that there is no inducing input point $\ve{x_{u_i}}$ close to our new training point $\ve{\hat{x}_+^{n+1}}$. As a result, we add $\ve{\hat{x}_+^{n+1}}$ to our set of inducing input points. This guarantees that each training point is close to at least one inducing input point, which always allows the data from the measurement to be taken into account.

\subsection{Predictions for stochastic test points}\label{ss:StochasticTestPointPredictions}

For deterministic test points $\ve{x_*}$ we can simply make use of~\eqref{eq:SparseGPStep2} to compute predictions. However, for a stochastic test point $\ve{x_*} \sim \N\left(\ve{\hat{x}_*}, \Sigma_{*_x}\right)$ it is more challening, since we have to calculate the distribution of $\ve{f_*} = \ve{f}(\ve{x_*})$, requiring us to solve an integration problem. This will not result in a Gaussian distribution, so once more we will apply moment matching. Previously, we had to make additional assumptions, to make sure that the mean vector and the covariance matrix could be solved for analytically. This time we do not have to. \cite{DeisenrothThesis} showed, based on work by~\cite{UncertainInputArticleCandela,UncertainInputArticleGirard}, that for the squared exponential covariance function (and also various other functions) the mean vector and the covariance matrix can be calculated analytically. We can apply the same ideas in our present setting.

For our results, we will first define some helpful quantities. When doing so, we should note that in theory every output $f_k(\ve{x})$ can have its own covariance function $k_k(\ldots)$, and as such its own hyperparameters $\alpha_k$ and $\Lambda_k$. (See Section~\ref{ss:MultipleOutputs}.) Keeping this in mind, we now define the vectors $\ve{q^k}$ and matrices $Q^{kl}$ element-wise as
\begin{align}
q_i^k & = \int_X k_k(\ve{x_{u_i}}, \ve{x_*}) p(\ve{x_*}) \, d\ve{x_*} \nonumber \\
& = \frac{\alpha_k^2}{\sqrt{|\Sigma_{*_x}||\Sigma_{*_x}^{-1} + \Lambda_k^{-1}|}} \exp\left(-\frac{1}{2} \left(\ve{x_{u_i}} - \ve{\ve{\hat{x}_*}}\right)^T \left(\Lambda_k + \Sigma_{*_x}\right)^{-1} \left(\ve{x_{u_i}} - \ve{\ve{\hat{x}_*}}\right)\right), \\
Q_{ij}^{kl} & = \int_X k_k(\ve{x_{u_i}}, \ve{x_*}) k_l(\ve{x_*}, \ve{x_{u_j}}) p(\ve{x_*}) \, d\ve{x_*} \nonumber \\
& = \frac{\alpha_k^2 \alpha_l^2}{\sqrt{|\Sigma_{*_x}||\Sigma_{*_x}^{-1} + \Lambda_k^{-1} + \Lambda_l^{-1}|}} \exp\left(-\frac{1}{2} \left(\ve{x_{u_i}} - \ve{x_{u_j}}\right)^T \left(\Lambda_k + \Lambda_l\right)^{-1} \left(\ve{x_{u_i}} - \ve{x_{u_j}}\right)\right) \nonumber \\
& \hspace{62pt} \exp\left(-\frac{1}{2} \left(\ve{\bar{x}_{u_{ij}}^{kl}} - \ve{\ve{\hat{x}_*}}\right)^T \left(\left(\Lambda_k^{-1} + \Lambda_l^{-1}\right)^{-1} + \Sigma_{*_x}\right)^{-1} \left(\ve{\bar{x}_{u_{ij}}^{kl}} - \ve{\ve{\hat{x}_*}}\right)\right),
\end{align}
where we have defined
\begin{equation}
\ve{\bar{x}_{u_{ij}}^{kl}} = \left(\Lambda_k^{-1} + \Lambda_l^{-1}\right)^{-1} \left(\Lambda_k^{-1} \ve{x_{u_i}} + \Lambda_l^{-1} \ve{x_{u_j}}\right).
\end{equation}
With these quantities, we can find that
\begin{align}
\ve{f_*} & \sim \N(\ve{\mu_*},\Sigma_*), \nonumber \\
\left[\ve{\mu_*}\right]_k & = \left(\ve{q^k}\right)^T \left(K_{uu}^k\right)^{-1} \ve{\mu_u^k}, \nonumber \\
\left[\Sigma_*\right]_{k,k} & = \alpha_k^2 - \tr\left(\left(K_{uu}^k\right)^{-1} \left(K_{uu}^k - \Sigma_u^k\right) \left(K_{uu}^k\right)^{-1} Q^{kk}\right) \nonumber \\
& \hspace{40pt} + \left(\ve{\mu_u^k}\right)^T \left(K_{uu}^k\right)^{-1} Q^{kk} \left(K_{uu}^k\right)^{-1} \ve{\mu_u^k} - \left[\ve{\mu_*}\right]_k^2, \nonumber \\
\left[\Sigma_*\right]_{k,l} & = \left(\ve{\mu_u^k}\right)^T \left(K_{uu}^k\right)^{-1} Q^{kl} \left(K_{uu}^l\right)^{-1} \ve{\mu_u^l} - \left[\ve{\mu_*}\right]_k \left[\ve{\mu_*}\right]_l,
\end{align}
where the latter expression is for the non-diagonal terms of $\Sigma_*$ (with $k \neq l$). Note that the first line of the above expression is in fact an approximation. In reality the distribution $\ve{f_*}$ is not Gaussian. The other two lines, however, are the analytic mean vector and covariance matrix. With these quantities, we can accurately predict the distribution of the output $\ve{f_*}$ for stochastic test points $\ve{x_*}$.

\begin{algorithm}[!htp]
	\label{alg:SISONIG}
	\SetKwBlock{Input}{Input:}{}
	\Input{
		A set of inputs $\ve{u_1}, \ve{u_2}, \ldots$ and outputs $\ve{y_1}, \ve{y_2}, \ldots$ of a system that is to be identified. Both the input and the output can be disturbed by noise.
	}
	\SetKwBlock{Prepare}{Preparation:}{}
	\Prepare{
		Define hyperparameters, either through the NIGP algorithm or by using expert knowledge about the system. Optionally, also define an initial set of inducing input points $X_u$.
	}
	\SetKwBlock{Update}{Updating:}{}
	\Update{
		\While{there are unprocessed measurements $\ve{y_{k+1}}$}{
			1. Set up $\ve{x_{k+1}}$ (shortened to $\ve{x_+}$) using its definition in~\eqref{eq:SystemFunction}. Find its prior distribution using known covariances between system outputs $\ve{y_k}, \ldots, \ve{y_{k-(n_y-1)}}$ and (if necessary) system inputs $\ve{u_k}, \ldots, \ve{u_{k-(n_u-1)}}$. Also find the prior distribution of the function output $\ve{y_{k+1}}$ (denoted as $\ve{f_{k+1}}$ or shortened as $\ve{f_+}$). \\
			2. Apply~\eqref{eq:PosteriorMeasurementPointDistribution} to find the posterior distribution $\N\left(\ve{\hat{x}_+^{k+1}},\Sigma_{+_x}^{k+1}\right)$ of $\ve{x_+}$. Optionally, use this to update the posterior distribution of the system outputs $\ve{y_k}, \ldots, \ve{y_{k-(n_y-1)}}$ and system inputs $\ve{u_k}, \ldots, \ve{u_{k-(n_u-1)}}$. \\
			3. Optionally, if $\ve{\hat{x}_+^{k+1}}$ is far removed from any inducing input point, add it to the set of inducing inputs $X_u$ using~\eqref{eq:AddingInducingInputPoints}. (Or rearrange/tune the inducing input points in any desired way.) \\
			4. Calculate the posterior distribution of the inducing input vector $\ve{f_u}$ for each of the outputs of $\ve{\phi}$ using~\eqref{eq:SONIGUpdate}. \\
			5. Calculate the posterior distribution of $\ve{y_{k+1}}$ using~\eqref{eq:PosteriorOutputDistribution}. Additionally, calculate the covariances between $\ve{y_{k+1}}$ and each of the previous system outputs $\ve{y_k}, \ldots, \ve{y_{k-(n_y-1)}}$ and inputs $\ve{u_k}, \ldots, \ve{u_{k-(n_u-1)}}$ through~\eqref{eq:PosteriorOutputCovariance}.
		}
	}
	\SetKwBlock{Predict}{Prediction:}{}
	\Predict{
		For any deterministic set of previous outputs $\ve{y_k}, \ldots, \ve{y_{k-(n_y-1)}}$ and inputs $\ve{u_k}, \ldots, \ve{u_{k-(n_u-1)}}$, apply~\eqref{eq:SparseGPStep2} to predict the next output $\ve{y_{k+1}}$. For stochastic outputs and inputs, use the expansion from Section~\ref{ss:StochasticTestPointPredictions}.
	}
	\hspace{1pt}\\ 
	\hspace{1pt}\\ 
	\caption{The steps required to identify nonlinear systems with measurement noise in an online way using the SONIG method.}
\end{algorithm}

\section{Experimental results}\label{s:TestResults}

In this section we apply the developed algorithm to test problems and compare its performance to existing state of the art solutions. First we apply the basic SONIG algorithm (Algorithm~\ref{alg:SONIG}) to approximate a sample function, allowing us to compare its performance to other regression algorithms. The results of this are discussed in Section~\ref{ss:SonigTest}. Then we apply the SONIG algorithm with all the extensions from Section~\ref{s:SonigDetails} (Algorithm~\ref{alg:SISONIG}) to identify a magneto-rheological fluid damper, the outcome of which is reported in Section~\ref{ss:MethodApplication}. All code for these examples, as well as for using the SONIG algorithm in general, is available on GitHub, see~\cite{SONIGCode}.

\subsection{Evaluating the SONIG algorithm through a sample function} \label{ss:SonigTest}

To compare the SONIG algorithm with other algorithms, we have set up a basic single-input single-output GP experiment. First, we randomly generate a sample function from a Gaussian process with a squared exponential covariance function (see~\eqref{eq:SECovarianceFunction}). This is done on the range $x \in [-5,5]$, subject to the hyperparameters $\alpha = 1$ and $\Lambda = 1$. Subsequently, we take $n$ training points at random places in the input range and distort both the input $x$ and the output $y$ with zero-mean Gaussian white noise with standard deviation $\sigma_x = 0.4$ and $\sigma_n = 0.1$, respectively. We use $n = 200$ unless mentioned otherwise. To this data set, we then apply the following algorithms.

\begin{enumerate}[(1)]
	\item GP regression without any input noise and with the exact hyperparameters, given above. This serves as a reference case: all other algorithms get noisy input points and tuned hyperparameters.
	\item GP regression with input noise and with hyperparameters tuned through the maximum-likelihood method.
	\item The NIGP algorithm of~\cite{GPWithInputNoise}. This algorithm has its own method of tuning hyperparameters, including $\sigma_x$.
	\item The SONIG algorithm, starting with $\ve{\mu_u^0} = \ve{m_u}$ and $\Sigma_{uu}^0 = K_{uu}$, using the hyperparameters given by (3). We use $X_u = \{-5, -4.5, -4, \ldots, 5\}$, resulting in $n_u = 21$ evenly distributed inducing input points.
	\item The same as (4), but now with more training points (800 instead of 200). Because the SONIG algorithm is computationally more efficient than the NIGP algorithm, the runtime of this is similar to that of (3), being roughly 2-3 seconds when using Matlab, although this of course does depend on the exact implementation of the algorithms.
	\item NIGP applied on a subset of data (100 training points) to predict the distribution of the inducing input points, followed by the SONIG algorithm applied to the remainder (700) of the training set, further updating the inducing input points. The runtime of this approach is again similar to that of (3), being 2-3 seconds.
	\item The FITC algorithm, using the hyperparameters of (2). This serves as a reference case.
\end{enumerate}

For all these algorithms, we examine both the Mean Squared Error (MSE) of the resulting prediction and the mean variance given by the regression algorithm. The latter is basically the estimate by the regression algorithm of the MSE. By comparing it with the real MSE, we learn about the integrity of the algorithm. As such, the ratio between these two is an indication of the algorithm integrity. We do this whole process $400$ times, each time for a different randomly generated sample function from the Gaussian process. The average of the results is subsequently shown in Table~\ref{t:ExperimentResults}.

\begin{table}[!t]
	\begin{center}
		\begin{footnotesize}
			\setlength{\tabcolsep}{.3em}
			\renewcommand{\arraystretch}{1.1}
			\caption{Comparison of various GP regression algorithms, applied to noisy measurements of $400$ randomly generated sample functions. For details, see the main text.}
			\label{t:ExperimentResults}
			\begin{tabular}{lcccc}
				\toprule
				 & $n$ & MSE & Mean variance & Ratio \\
				\midrule
				(1) GPR with exact hyperparameters and no input noise & $200$ & $0.87 \cdot 10^{-3}$ & $0.85 \cdot 10^{-3}$ & $1.02$ \\
				\hline
				(2) GPR with tuned hyperparameters & $200$ & $28.0 \cdot 10^{-3}$ & $8.3 \cdot 10^{-3}$ & $3.4$ \\
				\hline
				(3) NIGP with its own hyperparameter tuning & $200$ & $26.2 \cdot 10^{-3}$ & $5.6 \cdot 10^{-3}$ & $4.7$ \\
				\hline
				(4) SONIG using the hyperparameters of (3) & $200$ & $21.5 \cdot 10^{-3}$ & $8.1 \cdot 10^{-3}$ & $2.7$ \\
				\hline
				(5) SONIG using the hyperparameters of (3) & $800$ & $12.5 \cdot 10^{-3}$ & $2.2 \cdot 10^{-3}$ & $5.6$ \\
				\hline
				(6) NIGP on a subset, followed by SONIG on the rest & $100/700$ & $16.5 \cdot 10^{-3}$ & $2.3 \cdot 10^{-3}$ & $7.1$ \\
				\hline
				(7) FITC, using the hyperparameters of (2) & $800$ & $19.5 \cdot 10^{-3}$ & $2.7 \cdot 10^{-3}$ & $7.1$ \\
				\bottomrule
			\end{tabular}
		\end{footnotesize}
	\end{center}
\end{table}

There are many things that can be noticed from Table~\ref{t:ExperimentResults}. First of all, it is that for the given type of functions, and for an equal number of training points, the SONIG algorithm performs better than the NIGP algorithm. This is surprising, because the SONIG algorithm can be seen as a computationally efficient approximation of the NIGP algorithm. Further experiments have shown that this is mostly because the SONIG term takes into account the second derivative of the mean in its approximation; see $\mu_{u_i}^{n+1}$ from~\eqref{eq:SONIGUpdate}. The NIGP algorithm does not, and if SONIG also does not (detailed experiment results not included here for sake of brevity) the performance of the two algorithms is comparable.

A second thing that can be noticed is that more training points provide a higher accuracy. In particular, even the FITC algorithm (which does not take input noise into account) with 800 training points performs better than the NIGP or SONIG algorithms with 200 training points. It should be noted here that part of the reason is the type of function used: for functions with a steeper slope, it is expected that the NIGP and SONIG algorithms still perform better than FITC.

Finally, it is interesting to note that all algorithms, with the exception of regular GP regression with the exact hyperparameters, are much more optimistic about their predictions than is reasonable. That is, the ratio between the MSE and the mean variance is way larger than the value of $1$ which it should have. Ideally, the predicted variance of all algorithms would be significantly higher.

\begin{figure}
	\centering
	\begin{minipage}{.5\textwidth}
		\raggedright
		\includegraphics[width=.95\linewidth]{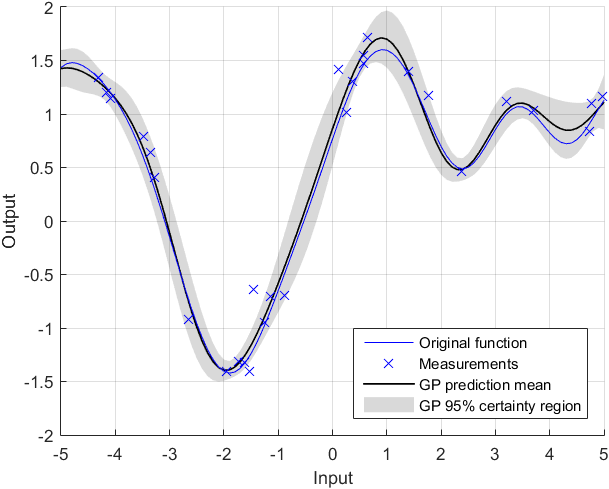}
	\end{minipage}%
	\begin{minipage}{.5\textwidth}
		\raggedleft
		\includegraphics[width=.95\linewidth]{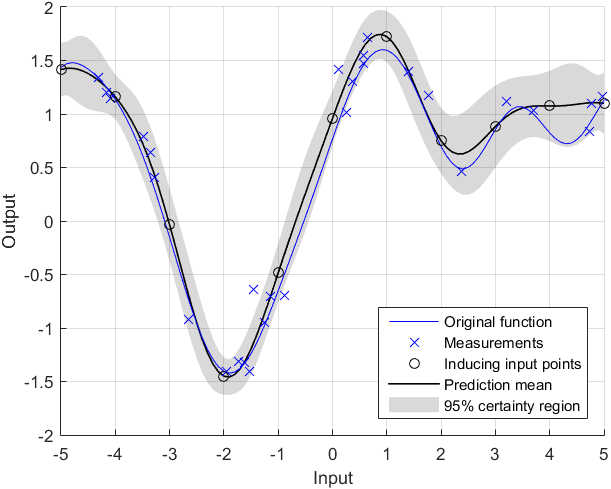}
	\end{minipage}
	\caption{Predictions of the NIGP algorithm (left) and the SONIG algorithm (right) after $n = 30$ training points have been incorporated. Exact conditions are described in the main text.}
	\label{fig:MethodComparison}
\end{figure}

Next, we will look at some plots. To be precise, we will examine algorithms (3) and (4) closer, but subject to only $n = 30$ training points and with $X_u = \{-5, -4, -3, \ldots, 5\}$, giving us $n_u = 11$ inducing input points. The predictions of the two algorithms, for a single random sample function, are shown in Figure~\ref{fig:MethodComparison}.

The most important thing that can be noticed here is that (for both methods) the posterior standard deviation varies with the slope of the to-be-approximated function. When the input is near $-2$, and the function is nearly flat, the standard deviation is small (well below $0.1$). However, when the input is near $-3$ or $-1/2$, the standard deviation is larger (near $0.2$). This is what can be expected, because measurements in these steep regions are much more affected/distorted by the noise, and hence provide less information.

A second thing to be noticed is the difference between the two methods. Especially for $x > 2$, where there are relatively few training points, the SONIG algorithm gives much higher variances. There are two reasons for this. The first is inherent to sparse algorithms. (The FITC algorithm would show a similar trend.) The second reason is inherent to the SONIG algorithm. Whereas regular GP regression (and similarly the NIGP algorithm) uses all training points together, the SONIG algorithm only uses data from previous training points while incorporating a new training point. As a result, when there are relatively few measurements in a certain region, and many of these measurements appear early in the updating process, the accuracy in that region can be expected to be slightly lower. However, as more training points are incorporated, which can be done very efficiently, the problem will quickly disappear.

\subsection{Identifying the dynamics of a magneto-rheological fluid damper} \label{ss:MethodApplication}

In the next experiment we will apply the developed system identification algorithm (Algorithm~\ref{alg:SISONIG}) to a practical problem. In particular, we model the dynamical behavior of a magneto-rheological fluid damper. The measured data for this example was provided by~\cite{FluidDamperData} and supplied through~\cite{DamperToolbox}, which also discusses various system identification examples using the techniques from~\cite{SIBookLjung}. This example is a common benchmark in system identification applications. It has for instance been used more recently in the context of Gaussian Process State Space Models (GP-SSM) by~\cite{SvenssonS:2017} in their Reduced Rank GP-SSM (RR GP-SSM) algorithm.

This example has 3499 measurements provided, sampled every $\Delta t = 0.05$ seconds. We will use the first 2000 measurements (10 seconds) for training and the next 1499 measurements (7.5 seconds) for evaluation. \cite{DamperToolbox} recommended to use one past output and three past inputs to predict subsequent outputs. Based on this, we learn a black-box model of the following functional form
\begin{equation}
y_{k+1} = \phi(y_k, u_k, u_{k-1}, u_{k-2}).
\end{equation}
Hyperparameters were tuned by passing a subset of the data to the NIGP algorithm. Rounded off for simplicity (which did not affect performance) they equaled
\begin{align}
\Lambda & = \diag\left(70^2, 20^2, 10^2, 10^2\right), & \alpha^2 & = 70^2, \nonumber \\
\Sigma_{+_x} & = \diag\left(2^2, 0.1^2, 0.1^2, 0.1^2\right), & \Sigma_{+_f} & = 2^2.
\end{align}
After processing a measurement $y_{k+1}$, the SONIG algorithm provided us with a posterior distribution of $y_{k+1}$, $y_k$, $u_k$, $u_{k-1}$ and $u_{k-2}$. The marginal posterior distribution of $y_{k+1}$, $u_k$ and $u_{k-1}$ was then used as prior distribution while incorporating the next measurement. Inducing input points were added online, as specified in Section~\ref{ss:AddingInducingInputPoints}, which eventually gave us $32$ inducing input points. This is a low number, and as a result, the whole training was done in only a few (roughly 10) seconds. As a result, we did not need to remove or shift inducing input points in any way.

\begin{figure}[!thp]
	\centering
	\includegraphics[width=0.8\textwidth]{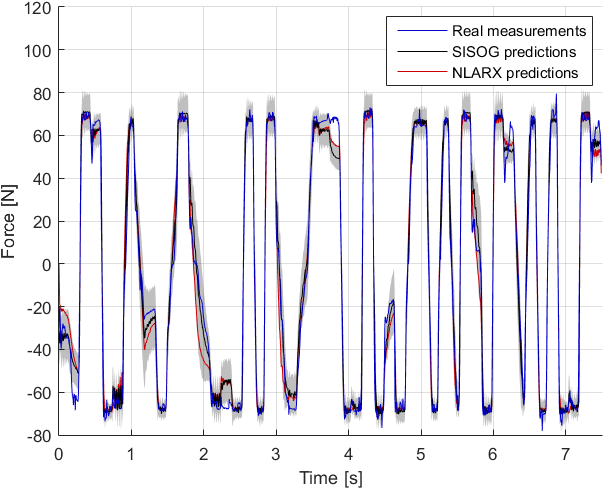}
	\caption{Prediction of the output of the magneto-rheological fluid damper by the SONIG algorithm (black) compared to the real output (blue). The grey area represents the 95\% uncertainty region as given by the algorithm. It shows that in the transition regions (like near $t = 2\thinspace$s) which the algorithm is less well trained on, the uncertainty is larger. It also shows (for instance near $t = 4.6\thinspace$s) that the uncertainty may grow over time. This is a result of the algorithm taking into account its own uncertainty in making future predictions. As comparison, also the best nonlinear ARX model predictions from~\cite{DamperToolbox} (red) are plotted. It is interesting to note that this model makes very similar errors as the SONIG algorithm, indicating the errors are mostly caused by distortions in the training/evaluation data.}
	\label{fig:OutputPredictions}
\end{figure}

After all training measurements had been used, the SONIG algorithm was given the input data for the remaining 1499 measurements, but not the output data. It had to predict this output data by itself, using each prediction $y_k$ to predict the subsequent $y_{k+1}$. While doing so, the algorithm also calculated the variance of each prediction $y_k$, taking this into account while predicting the next output using the techniques from Section~\ref{ss:StochasticTestPointPredictions}. The resulting predictions can be seen in Figure~\ref{fig:OutputPredictions}.

A comparison of the algorithm with various other methods is shown in Table~\ref{t:AlgorithmComparison}. We also added in regular GP regression and NIGP, applied to the ARX model~\eqref{eq:SystemFunction}, as comparison. This table shows that the SONIG algorithm, when applied in its system identification set-up, can clearly outperform other black-box modeling approaches. It is better than regular GP regression at taking into account uncertainties and better than NIGP mainly due to the reasons explained before. It should be noted here, however, that this is all subject to the proper tuning of hyperparameters and the proper choice of inducing input points. With different hyperparameters or inducing input point selection strategies, the performance of the SONIG algorithm will degrade slightly, though it is still likely to outperform other identification algorithms.

\begin{table}[!htp]
	\begin{center}
		\begin{footnotesize}
			\setlength{\tabcolsep}{.3em}
			\renewcommand{\arraystretch}{1.1}
			\caption{Comparison of various system identification models and algorithms when applied to data from the magneto-rheological fluid damper. All algorithms were given 2000 measurements for training and 1499 measurements for evaluation.}
			\label{t:AlgorithmComparison}
			\begin{tabular}{lcr}
				\toprule
				\textbf{Algorithm} & \hspace{20pt} \textbf{RMSE} \hspace{20pt} & \textbf{Source} \\
				\midrule
				Linear OE model (4th order) & 27.1 & \cite{DamperToolbox} \\
				\hline
				Hammerstein-Wiener (4th order) & 27.0 & \cite{DamperToolbox} \\
				\hline
				NLARX (3rd order, wavelet network) & 24.5 & \cite{DamperToolbox} \\
				\hline
				NLARX (3rd order, tree partition) & 19.3 & \cite{DamperToolbox} \\
				\hline
				NIGP & 10.2 & This paper \\
				\hline
				GP regression & 9.87 & This paper \\
				\hline
				NLARX (3rd order, sigmoid network) & 8.24 & \cite{DamperToolbox} \\
				\hline
				RR GP-SSM & 8.17 & \cite{GPSSMApplication} \\
				\hline
				\textbf{SONIG} & \textbf{7.12} & This paper \\
				\bottomrule
			\end{tabular}
		\end{footnotesize}
	\end{center}
\end{table}

\section{Conclusions and recommendations}\label{s:Conclusions}

We can conclude that the presented SONIG algorithm works as intended. Just like the FITC algorithm that it expands upon, it is mainly effective when there are more measurements than the NIGP algorithm (or regular GP regression) can handle. The SONIG algorithm can then include the additional measurements very efficiently---incorporating each training point in constant runtime---resulting in a higher accuracy than what the NIGP algorithm could have achieved. However, even when this is not the case, the SONIG algorithm has on average a better performance than the NIGP algorithm, though it still needs the NIGP algorithm for hyperparameter tuning.

Though the SONIG algorithm can be used for any type of regression problem, it has been successfully applied, in its system identification set-up, to a nonlinear black-box system identification problem. With the proper choice of hyperparameters and inducing input points, it outperformed existing state-of-the-art nonlinear system identification algorithms.

Nevertheless, there are still many improvements that can be made to the SONIG algorithm. For instance, to improve the accuracy of the algorithm, we can look at reducing some of the approximating assumptions, like the linearization assumption~\eqref{eq:LinearizedGPAssumption} or the assumption that higher order terms of $\Sigma_+$ are negligible.

Another way to improve the accuracy of the algorithm is to increase the number of inducing input points, but this will slow down the algorithm. To compensate, we could look into updating only the few nearest inducing input points (with the highest covariance) when incorporating a new training point. Experience has shown that updates hardly affect inducing inputs far away from the training point (with a low covariance) so this could lead to more efficient updates.

A final possible improvement would concern the addition of a smoothing step in the algorithm. Currently, early measurements are used to provide more accuracy for later measurements, but not vice versa. If we also walk back through the measurements, like in a smoothing algorithm, a higher accuracy might be obtained. 

\acks{This research is supported by the Dutch Technology Foundation STW, which is part of the Netherlands Organisation for Scientific  Research (NWO), and which is partly funded by the Ministry of Economic Affairs. The work was also supported by the Swedish research Council (VR) via the projects \emph{NewLEADS - New Directions in Learning Dynamical Systems} and \emph{Probabilistic modeling of dynamical systems} (Contract number: 621-2016-06079, 621-2013-5524) and by the Swedish Foundation for Strategic Research (SSF) via the project \emph{ASSEMBLE} (Contract number: RIT15-0012). We would also like to thank Marc Deisenroth for fruitful discussion and in particular for pointing us to the NIGP algorithm.}

\appendix
\section*{Appendix. Derivatives of prediction matrices}\label{s:PredictionMatricesDerivatives}

The SONIG update law~\eqref{eq:SONIGUpdate} contains various derivatives of matrices. Using~\eqref{eq:SparseOnlineGPPrediction} and~\eqref{eq:SparseOnlineGPUpdate} we can find them. To do so, we first define the scalar quantity
\begin{equation}
P = \Sigma_{++}^n + \sigma_n^2 = K_{++} + \sigma_n^2 - K_{+u} K_{uu}^{-1} \left(K_{uu} - \Sigma_{uu}^n\right) K_{uu}^{-1} K_{u+}.
\end{equation}
We also assume that $m(\ve{x}) = 0$ for ease of notation. (If not, this can of course be taken into account.) The derivatives of $\ve{\mu_u^{n+1}}$ and $\Sigma_{uu}^{n+1}$ can now be found element-wise through
\begin{align}
\hspace{-10pt} \frac{d\ve{\mu_u^{n+1}}}{dx_{+_j}} & = \Sigma_{uu}^n K_{uu}^{-1} \left(\frac{dK_{u+}}{dx_{+_j}} P^{-1} \left(y_+ - K_{+u} K_{uu}^{-1} \ve{\mu_u^n}\right) + K_{u+} \frac{dP^{-1}}{dx_{+_j}} \left(y_+ - K_{+u} K_{uu}^{-1} \ve{\mu_u^n}\right)\right. \nonumber \\
& \hspace{12pt} - \left. K_{u+} P^{-1} \frac{dK_{+u}}{dx_{+_j}} K_{uu}^{-1} \ve{\mu_u^n}\right), \nonumber \\
\hspace{-10pt} \frac{d^2\ve{\mu_u^{n+1}}}{dx_{+_j}dx_{+_k}} & = \Sigma_{uu}^n K_{uu}^{-1} \left(\frac{d^2K_{u+}}{dx_{+_j}dx_{+_k}} P^{-1} \left(y_+ - K_{+u} K_{uu}^{-1} \ve{\mu_u^n}\right) + \frac{dK_{u+}}{dx_{+_j}} \frac{dP^{-1}}{dx_{+_k}} \left(y_+ - K_{+u} K_{uu}^{-1} \ve{\mu_u^n}\right) \right. \nonumber \\
& \hspace{12pt} \left. - \frac{dK_{u+}}{dx_{+_j}} P^{-1} \frac{dK_{+u}}{dx_{+_k}} K_{uu}^{-1} \ve{\mu_u^n} + \frac{dK_{u+}}{dx_{+_k}} \frac{dP^{-1}}{dx_{+_j}} \left(y_+ - K_{+u} K_{uu}^{-1} \ve{\mu_u^n}\right) \right. \nonumber \\
& \hspace{12pt} \left. + K_{u+} \frac{d^2P^{-1}}{dx_{+_j}dx_{+_k}} \left(y_+ - K_{+u} K_{uu}^{-1} \ve{\mu_u^n}\right) - K_{u+} \frac{dP^{-1}}{dx_{+_j}} \frac{dK_{+u}}{dx_{+_k}} K_{uu}^{-1} \ve{\mu_u^n} \right. \nonumber \\
& \hspace{12pt}  \left. - \frac{dK_{u+}}{dx_{+_k}} P^{-1} \frac{dK_{+u}}{dx_{+_j}} K_{uu}^{-1} \ve{\mu_u^n} - K_{u+} \frac{dP^{-1}}{dx_{+_k}} \frac{dK_{+u}}{dx_{+_j}} K_{uu}^{-1} \ve{\mu_u^n} - K_{u+} P^{-1} \frac{d^2K_{+u}}{dx_{+_j}dx_{+_k}} K_{uu}^{-1} \ve{\mu_u^n}\right), \nonumber \\
\hspace{-10pt} \frac{d\Sigma_{uu}^{n+1}}{dx_{+_j}} & = -\Sigma_{uu}^n K_{uu}^{-1} \left(\frac{dK_{u+}}{dx_{+_j}} P^{-1} K_{+u} + K_{u+} \frac{dP^{-1}}{dx_{+_j}} K_{+u} + K_{u+} P^{-1} \frac{dK_{+u}}{dx_{+_j}}\right) K_{uu}^{-1} \Sigma_{uu}^n, \nonumber \\
\hspace{-10pt} \frac{d^2\Sigma_{uu}^{n+1}}{dx_{+_j}dx_{+_k}} & = -\Sigma_{uu}^n K_{uu}^{-1} \left(\frac{d^2K_{u+}}{dx_{+_j}dx_{+_k}} P^{-1} K_{+u} + \frac{dK_{u+}}{dx_{+_j}} \frac{dP^{-1}}{dx_{+_k}} K_{+u} + \frac{dK_{u+}}{dx_{+_j}} P^{-1} \frac{dK_{+u}}{dx_{+_k}}\right. \nonumber \\
& \hspace{20pt} \left. + \frac{dK_{u+}}{dx_{+_k}} \frac{dP^{-1}}{dx_{+_j}} K_{+u} + K_{u+} \frac{d^2P^{-1}}{dx_{+_j}dx_{+_k}} K_{+u} + K_{u+} \frac{dP^{-1}}{dx_{+_j}} \frac{dK_{+u}}{dx_{+_k}} \right. \nonumber \\
& \hspace{20pt} \left. + \frac{dK_{u+}}{dx_{+_k}} P^{-1} \frac{dK_{+u}}{dx_{+_j}} + K_{u+} \frac{dP^{-1}}{dx_{+_k}} \frac{dK_{+u}}{dx_{+_j}} + K_{u+} P^{-1} \frac{d^2K_{+u}}{dx_{+_j}dx_{+_k}}\right) K_{uu}^{-1} \Sigma_{uu}^n.
\end{align}
These expressions contain various additional derivatives. To find them, we need to choose a covariance function. (The above expressions are valid for any covariance function.) If we use the squared exponential covariance function of~\eqref{eq:SECovarianceFunction}, we can derive
\begin{align}
\hspace{-6pt} \frac{dK_{u_i+}}{d\ve{x_+}} & = \alpha^2 \exp\left(-\frac{1}{2} \left(\ve{x_{u_i}} - \ve{x_+}\right)^T \Lambda^{-1} \left(\ve{x_{u_i}} - \ve{x_+}\right)\right) \left(\ve{x_{u_i}} - \ve{x_+}\right)^T \Lambda^{-1}, \nonumber \\
\hspace{-6pt} \frac{d^2K_{u_i+}}{d\ve{x}_+^2} & = \alpha^2 \exp\left(-\frac{1}{2} \left(\ve{x_{u_i}} - \ve{x_+}\right)^T \Lambda^{-1} \left(\ve{x_{u_i}} - \ve{x_+}\right)\right) \left(\Lambda^{-1} \left(\ve{x_{u_i}} - \ve{x_+}\right) \left(\ve{x_{u_i}} - \ve{x_+}\right)^T \Lambda^{-1} - \Lambda^{-1}\right), \nonumber \\
\hspace{-6pt} \frac{dP^{-1}}{d\ve{x_+}} & = -P^{-2} \frac{dP}{d\ve{x_+}} = 2P^{-2} \left(K_{+u} K_{uu}^{-1} (K_{uu} - \Sigma_{uu}^n) K_{uu}^{-1} \frac{dK_{u+}}{d\ve{x_+}}\right), \nonumber \\
\hspace{-6pt} \frac{d^2 P^{-1}}{d\ve{x}_+^2} & = \frac{d}{d\ve{x_+}}\left(-P^{-2} \frac{dP}{d\ve{x_+}}\right) = 2P^{-3} \left(\frac{dP}{d\ve{x_+}}\right)^T\left(\frac{dP}{d\ve{x_+}}\right) - P^{-2} \frac{d^2P}{d\ve{x}_+^2}, \nonumber \\
\hspace{-6pt} \frac{dP}{d\ve{x_+}} & = -2K_{+u} K_{uu}^{-1} (K_{uu} - \Sigma_{uu}^n) K_{uu}^{-1} \frac{dK_{u+}}{d\ve{x_+}}, \nonumber \\
\hspace{-6pt} \frac{d^2P}{d\ve{x}_+^2} & = -2\frac{dK_{+u}}{d\ve{x_+}} K_{uu}^{-1} (K_{uu} - \Sigma_{uu}^n) K_{uu}^{-1} \frac{dK_{u+}}{d\ve{x_+}} - 2K_{+u} K_{uu}^{-1} (K_{uu} - \Sigma_{uu}^n) K_{uu}^{-1} \frac{d^2K_{u+}}{d\ve{x}_+^2}.
\end{align}

\vskip 0.2in
\bibliography{bibliography}

\end{document}